\documentclass[12pt]{article}

\usepackage{amsmath,amsthm,amssymb, listings, tikz, pgf}
\usepackage[top=1.25in, bottom=1.25in, left=1.25in, right=1.25in]{geometry}
\usepackage{pgf}
\usepackage{tikz}

\usepackage{multicol}
\usepackage{enumitem}
\usepackage{sectsty}
\subsectionfont{\normalsize}
\subsubsectionfont{\small}
\usepackage{listings}

\usepackage{csquotes}
\usepackage{gensymb}
\usepackage[font=footnotesize]{caption}

\usetikzlibrary{arrows,automata}
\usepackage[latin1]{inputenc}

\usepackage[hidelinks]{hyperref}
\hypersetup{
    colorlinks,
    linkcolor={red!50!black},
    citecolor={blue!50!black},
    urlcolor={blue!80!black}
}

\colorlet{punct}{red!60!black}
\definecolor{background}{HTML}{EEEEEE}
\definecolor{delim}{RGB}{20,105,176}
\colorlet{numb}{magenta!60!black}
\lstdefinelanguage{json}{
    basicstyle=\normalfont\ttfamily,
    numbers=left,
    numberstyle=\scriptsize,
    stepnumber=1,
    numbersep=8pt,
    showstringspaces=false,
    breaklines=true,
    frame=lines,
    backgroundcolor=\color{background},
    literate=
     *{:}{{{\color{punct}{:}}}}{1}
      {,}{{{\color{punct}{,}}}}{1}
      {\{}{{{\color{delim}{\{}}}}{1}
      {\}}{{{\color{delim}{\}}}}}{1}
      {[}{{{\color{delim}{[}}}}{1}
      {]}{{{\color{delim}{]}}}}{1},
}

 \graphicspath{{figures/}}

\author{Kaiyu Zheng}
\date{September 2, 2016\footnote{Update on April 8th, 2019 (added information on \texttt{amcl})}}
\title{ROS Navigation Tuning Guide}

\begin{document}
\maketitle

\section*{Abstract}
The ROS navigation stack is powerful for mobile robots to move from place to place reliably. The job of navigation stack is to produce a safe path for the robot to execute, by processing data from odometry, sensors and environment map. Maximizing the performance of this navigation stack requires some fine tuning of parameters, and this is not as simple as it looks. One who is sophomoric about the concepts and reasoning may try things randomly, and wastes a lot of time.

This article intends to guide the reader through the process of fine tuning navigation parameters. It is the reference when someone need to know the "how" and "why" when setting the value of key parameters. This guide assumes that the reader has already set up the navigation stack and ready to optimize it. This is also a summary of my work with the ROS navigation stack.\\

\begin{multicols}{2}
[
\section*{Topics}
]
    \begin{flushleft}
        \begin{enumerate}
            \item Velocity and Acceleration\\
            \item Global Planner\\
            \begin{enumerate}
              \item Global Planner Selection
              \item Global Planner Parameters
            \end{enumerate}
            \item Local Planner\\
            \begin{enumerate}
              \item Local Planner Selection
              \item DWA Local Planner
                \begin{enumerate}
                    \item DWA algorithm
                    \item DWA forward simulation
                    \item DWA trajectory scoring
                    \item Other DWA parameters
                \end{enumerate}
            \end{enumerate}
            \item Costmap Parameters
            \item AMCL
            \item Recovery Behavior
            \item Dynamic Reconfigure
            \item Problems
        \end{enumerate}
    \end{flushleft}
\end{multicols}

\newpage

\section{Velocity and Acceleration}
This section concerns with synchro-drive robots. The dynamics (e.g. velocity and acceleration of the robot) of the robot is essential for local planners including dynamic window approach (DWA) and timed elastic band (TEB). In ROS navigation stack, local planner
takes in odometry messages ("odom" topic) and outputs velocity commands ("cmd\_vel" topic) that controls the robot's motion. \\

\noindent Max$/$min velocity and acceleration are two basic parameters for the mobile base. Setting them correctly is very helpful for optimal local planner behavior. In ROS navigation, we need to know translational and rotational velocity and acceleration.

\subsection{To obtain maximum velocity}
Usually you can refer to your mobile base's manual. For example, SCITOS G5 has maximum velocity 1.4 m$/$s\footnote{This information is obtained from \href{http://www.metralabs.com/en/research}{MetraLabs's website}.}. In ROS, you can also subscribe to the \texttt{odom} topic to obtain the current odometry information. If you can control your robot manually (e.g. with a joystick), you can try to run it forward until its
speed reaches constant, and then echo the odometry data.

\textit{Translational velocity} ($m/s$) is the velocity when robot is moving in a straight line. Its max value is the same as the maximum velocity we obtained above. \textit{Rotational velocity} ($rad/s$) is equivalent as angular velocity; its maximum value is the angular velocity of the robot when it is rotating in place. To obtain maximum rotational velocity, we can control the robot by a joystick and rotate the robot 360 degrees after the robot's speed reaches constant, and time this movement. 

For safety, we prefer to set maximum translational and rotational velocities to be lower than their actual maximum values.

\subsection{To obtain maximum acceleration}
There are many ways to measure maximum acceleration of your mobile base, if your manual does not tell you directly. 

In ROS, again we can echo odometry data which include time stamps, and them see how long it took the robot to reach constant maximum translational velocity ($t_{i}$). Then we use the position and velocity information from odometry (nav\_msgs/Odometry message) to compute the acceleration in this process. Do several trails and take the average. Use $t_t, t_r$ to denote the time used to reach translationand and rotational maximum velocity from static, respectively. The maximum translational acceleration $a_{t,max}=\text{max }dv / dt\approx v_{max}/t_t$. Likewise, rotational acceleration can be computed by $a_{r,max}=\text{max }d\omega / dt\approx \omega_{max}/t_r$.


\subsection{Setting minimum values}
Setting minimum velocity is not as formulaic as above. For minimum translational velocity, we want to set it to a large negative value
because this enables the robot to back off when it needs to unstuck itself, but it should prefer moving forward in most cases. For minimum rotational velocity, we also want to set it to negative (if the parameter allows) so that the robot can rotate in either directions. Notice that DWA Local Planner takes the absolute value of robot's minimum rotational velocity.

\subsection{Velocity in x, y direction}

\textbf{$x$ velocity} means the velocity in the direction parallel to robot's straight movement. It is the same as translational velocity. \textbf{$y$ velocity} is the velocity in the direction perpendicular to that straight movement. It is called "strafing velocity" in \texttt{teb\_local\_planner}. $y$ velocity should be set to zero for non-holonomic robot (such as
differential wheeled robots).\\

\section{Global Planner}

\subsection{Global Planner Selection}

To use the \texttt{move$\_$base} node in navigation stack, we need to have a global planner
and a local planner. There are three global planners that adhere to \texttt{nav\_core::BaseGlobal\\Planner} interface: \texttt{carrot$\_$planner},
\texttt{navfn} and \texttt{global$\_$planner}.

\subsubsection{carrot\_planner}

This is the simplest one. It checks if the given goal is an obstacle, and if so it picks an alternative goal close to the original one, by moving back along the vector between the robot and the goal point. Eventually it passes this valid goal as a plan to the local planner or controller (internally). Therefore, this planner does not do any global path planning. It is helpful if you require your robot to move close to the given goal even if the goal is unreachable. In complicated indoor environments, this planner is not very practical.

\subsubsection{navfn and global\_planner}

\texttt{navfn} uses Dijkstra's algorithm to find a global path with minimum cost between start point and
end point. \texttt{global$\_$planner} is built as a more flexible replacement of \texttt{navfn} with more options. These
options include (1) support for A$*$, (2) toggling quadratic approximation, (3) toggling grid path. Both \texttt{navfn} and
global planner are based on the work by \cite{brock1999high}:

\subsection{Global Planner Parameters}
\normalsize

Since \texttt{global\_planner} is generally the one that we prefer, let us look at some of its key parameters.
Note: not all of these parameters are listed on ROS's website, but you can see them if you run the rqt dynamic reconfigure program: with \begin{center}
\texttt{rosrun rqt\_reconfigure rqt\_reconfigure}
\end{center}

We can leave \texttt{allow\_unknown}(true), \texttt{use\_dijkstra}(true), \texttt{use\_quadratic}(true), \texttt{use\_grid\_path}(false), \texttt{old\_navfn\_behavior}(false) to
their default values. Setting \texttt{visualize\_potential}(false) to true is helpful when we would like to visualize the potential map in RVIZ.

\begin{figure}[!htb]
\minipage{0.49\textwidth}
  \includegraphics[width=\linewidth]{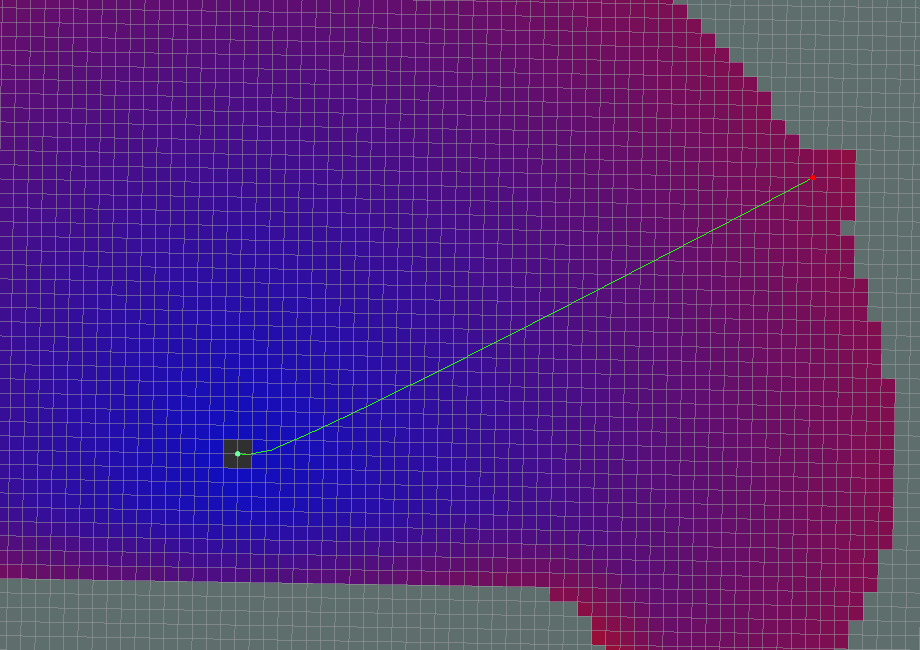}
  \caption{Dijkstra path}
\endminipage\hfill
\minipage{0.49\textwidth}
  \includegraphics[width=\linewidth]{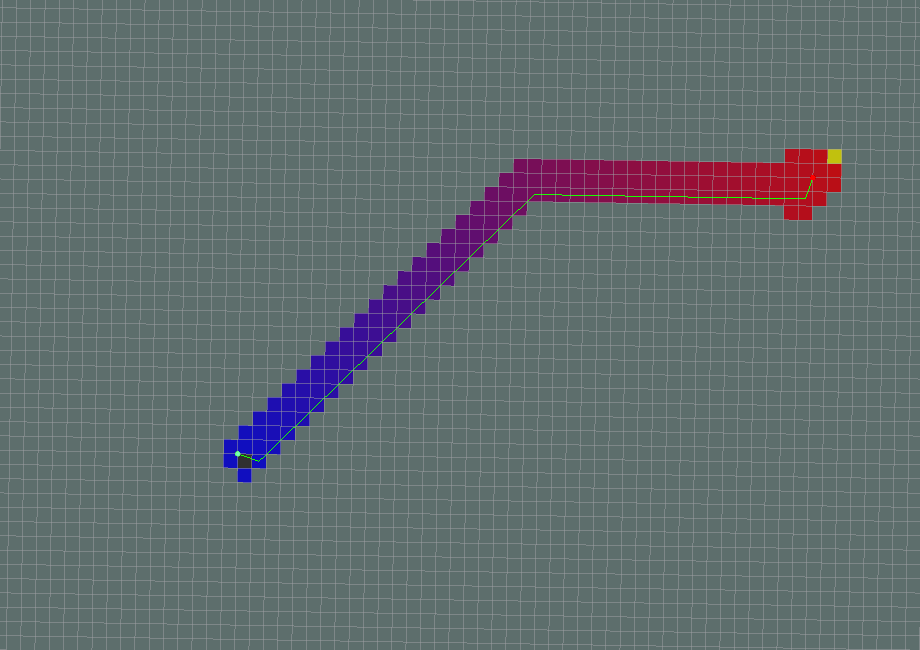}
  \caption{A* path}
\endminipage\hfill
\end{figure}

\begin{figure}[!htb]
\minipage{0.49\textwidth}
  \includegraphics[width=\linewidth]{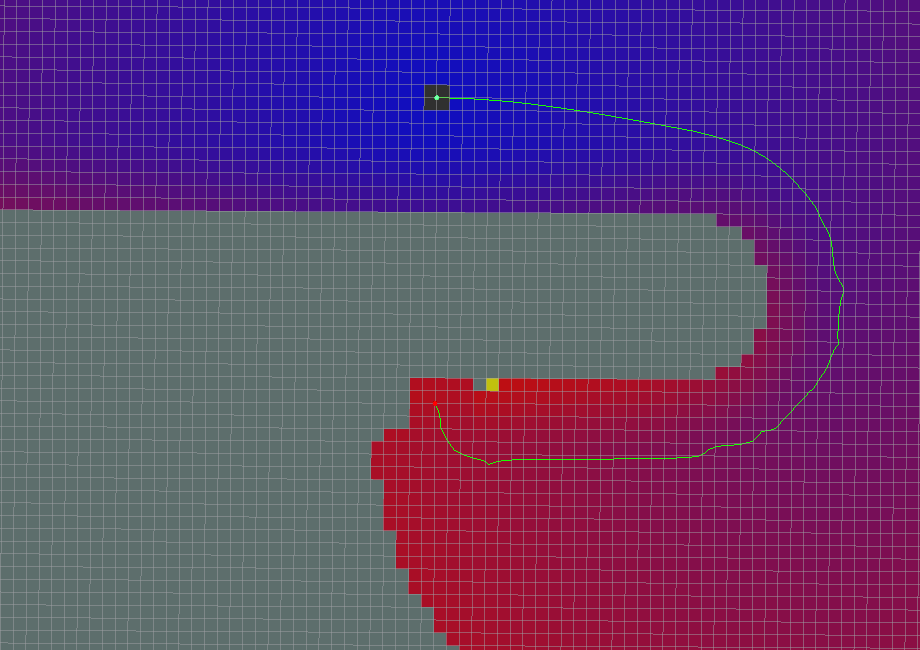}
  \caption{Standard Behavior}
\endminipage\hfill
\minipage{0.49\textwidth}
  \includegraphics[width=\linewidth]{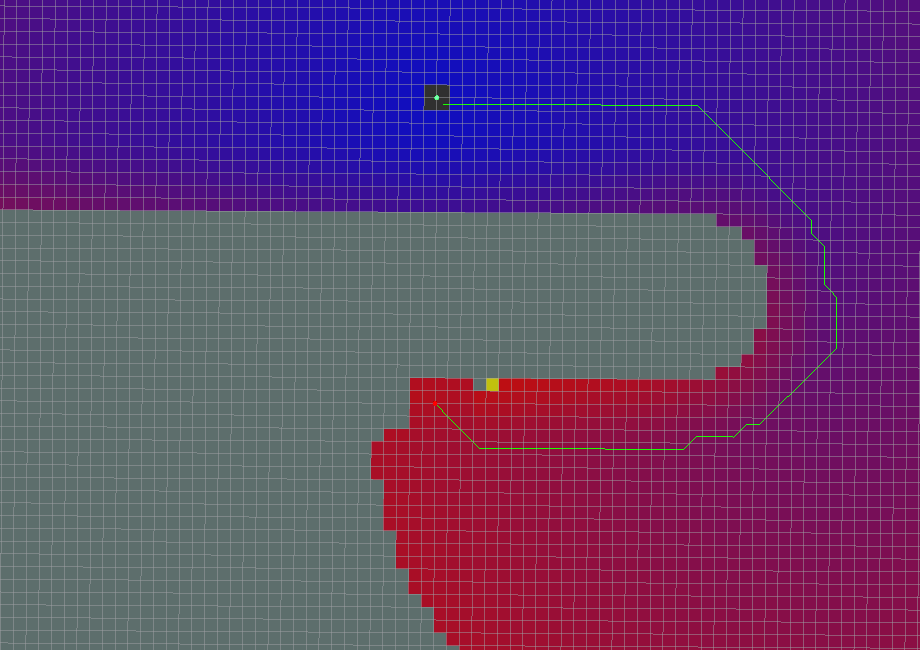}
  \caption{Grid Path}
\endminipage\hfill
\end{figure}

Besides these parameters, there are three other unlisted parameters that actually determine the quality of the planned
global path. They are \texttt{cost\_factor}, \texttt{neutral\_cost}, \texttt{lethal\_cost}. Actually, these parameters also present
in \texttt{navfn}. The source code\footnote{\scriptsize\url{https://github.com/ros-planning/navigation/blob/indigo-devel/navfn/include/navfn/navfn.h}} \normalsize has one paragraph explaining how \texttt{navfn} computes cost values.

\newpage

\begin{figure}[!htb]
\minipage{0.32\textwidth}
  \includegraphics[width=\linewidth]{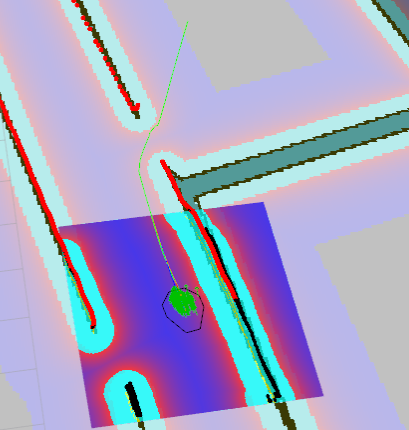}
  \caption{\texttt{cost\_factor} = 0.01}
\endminipage\hfill
\minipage{0.32\textwidth}
    \includegraphics[width=\linewidth]{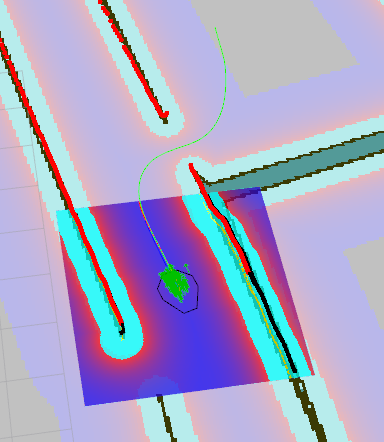}
  \caption{\texttt{cost\_factor} = 0.55}
\endminipage\hfill
\minipage{0.32\textwidth}
  \includegraphics[width=\linewidth]{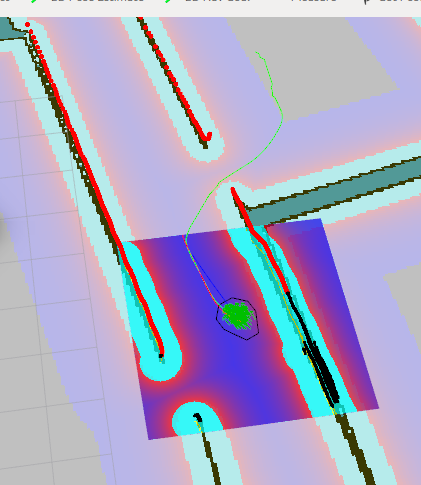}
  \caption{\texttt{cost\_factor} = 3.55}
\endminipage\hfill
\end{figure}

\begin{figure}[!htb]
\minipage{0.32\textwidth}
  \includegraphics[width=\linewidth]{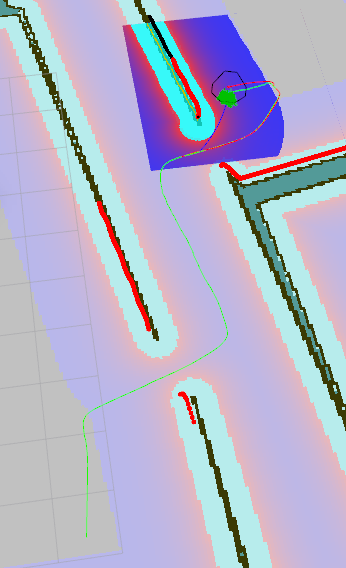}
  \caption{\texttt{neutral\_cost} = 1}
\endminipage\hfill
\minipage{0.32\textwidth}
  \includegraphics[width=\linewidth]{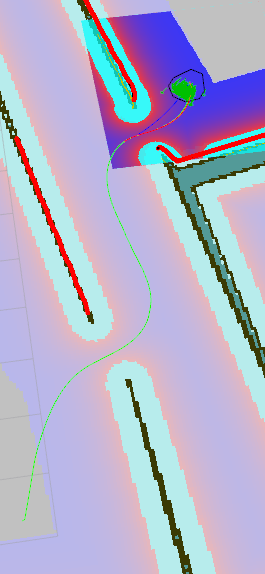}
  \caption{\texttt{neutral\_cost} = 66}
\endminipage\hfill
\minipage{0.32\textwidth}
  \includegraphics[width=\linewidth]{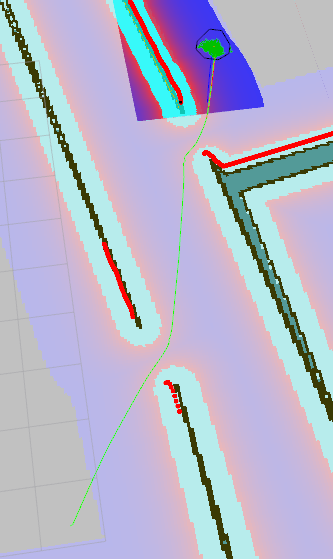}
  \caption{\texttt{neutral\_cost} = 233}
\endminipage\hfill
\end{figure}

\newpage

\texttt{navfn} cost values are set to

$$\texttt{cost = COST\_NEUTRAL + COST\_FACTOR * costmap\_cost\_value.}$$

Incoming costmap cost values are in the range 0 to 252. The comment also says:
\begin{displayquote}
    With \texttt{COST\_NEUTRAL} of 50, the \texttt{COST\_FACTOR} needs to be about 0.8 to
    ensure the input values are spread evenly over the output range, 50
    to 253.  If \texttt{COST\_FACTOR} is higher, cost values will have a plateau
    around obstacles and the planner will then treat (for example) the
    whole width of a narrow hallway as equally undesirable and thus
    will not plan paths down the center.
\end{displayquote}

\paragraph{Experiment observations} Experiments have confirmed this explanation. Setting \texttt{cost\_factor} to too low or too high lowers
the quality of the paths. These paths do not go through the middle of obstacles on each side and have relatively flat curvature. Extreme
\texttt{neutral\_cost} values have the same effect. For \texttt{lethal\_cost}, setting it to a low value may result
in failure to produce any path, even when a feasible path is obvious. Figures $5-10$ show the effect of \texttt{cost\_factor} and \texttt{neutral\_cost} on global path planning. The green line is the global path produced by \texttt{global\_planner}.

After a few experiments we observed that when \texttt{cost\_factor} = 0.55, \texttt{neutral\_cost = 66}, and \texttt{lethal\_cost = 253}, the global path is quite desirable.

\section{Local Planner Selection}

Local planners that adhere to \texttt{nav\_core::BaseLocalPlanner} interface are \texttt{dwa\_local\\\_planner}, \texttt{eband\_local\_planner} and \texttt{teb\_local\_planner}.
They use different algorithms to generate velocity commands. Usually \texttt{dwa\_local\_planner} is the go-to choice. We will discuss it in detail. More
information on other planners will be provided later.

\subsection{DWA Local Planner}
\subsubsection{DWA algorithm}

See next page.

\newpage
\texttt{dwa\_local\_planner} uses Dynamic Window Approach (DWA) algorithm. ROS Wiki provides a summary of its implementation of this algorithm:

\begin{center}
\setlength{\fboxsep}{1em}
\fbox{\begin{minipage}{33 em}
\begin{enumerate}
\item Discretely sample in the robot's control space \textit{(dx,dy,dtheta)}
    \item For each sampled velocity, perform forward simulation from the robot's current state to predict what would happen if the sampled velocity were applied for some (short) period of time.
    \item Evaluate (score) each trajectory resulting from the forward simulation, using a metric that incorporates characteristics such as: proximity to obstacles, proximity to the goal, proximity to the global path, and speed. Discard illegal trajectories (those that collide with obstacles).
    \item Pick the highest-scoring trajectory and send the associated velocity to the mobile base.
    \item Rinse and repeat.
\end{enumerate}
\end{minipage}}
\end{center}

\noindent DWA is proposed by \cite{fox1997dynamic}. According to this paper, the goal of DWA is to produce a $(v,\omega)$ pair which
represents a circular trajectory that is optimal for robot's local condition. DWA reaches this goal by searching the velocity space in the next time interval.
The velocities in this space are restricted to be admissible, which means the robot must be able to stop before reaching the closest obstacle on the
circular trajectory dictated by these admissible velocities. Also, DWA will only consider velocities within a dynamic window, which is defined to be the
set of velocity pairs that is reachable within the next time interval given the current translational and rotational velocities and accelerations. DWA maximizes an objective
function that depends on (1) the progress to the target, (2) clearance from obstacles, and (3) forward velocity to produce the optimal velocity pair.

Now, let us look at the algorithm summary on ROS Wiki. The first step is to sample velocity pairs $(v_x, v_y, \omega)$ in
the velocity space within the dynamic window. The second step is basically obliterating velocities (i.e. kill off bad trajectories) that are not admissible. The third step is
to evaluate the velocity pairs using the objective function, which outputs \textit{trajectory score}. The fourth and fifth steps are easy to understand: take the current best
velocity option and recompute.\\

This DWA planner depends on the local costmap which provides obstacle information. Therefore, tuning the parameters for the local costmap is crucial for optimal behavior of DWA local planner. Next, we will look at parameters in forward simulation, trajectory scoring, costmap, and so on.

\subsubsection{DWA Local Planner : Forward Simulation}

Forward simulation is the second step of the DWA algorithm. In this step, the local planner takes the velocity samples in robot's control space, and examine the circular trajectories represented by those velocity samples, and finally
eliminate bad velocities (ones whose trajectory intersects with an obstacle). Each velocity sample is simulated as if it is applied to the robot for a set time interval, controlled by \texttt{sim\_time}($s$)
parameter. We can think of \texttt{sim\_time} as the time allowed for the robot to move with the sampled velocities.

Through experiments, we observed that the longer the value of \texttt{sim\_time}, the heavier the computation load becomes. Also, when \texttt{sim\_time} gets
longer, the path produced by the local planner is longer as well, which is reasonable. Here are some suggestions on how to tune this \texttt{sim\_time} parameter.

\paragraph{How to tune \texttt{sim\_time}}
Setting \texttt{sim\_time} to a very low value ($<= 2.0$) will result in limited performance, especially when the robot needs to pass a narrow doorway, or gap between furnitures, because there is insufficient time to obtain the optimal trajectory that actually goes through the narrow passway. On the other hand, since with DWA Local Planner, all trajectories are simple arcs,
setting the \texttt{sim\_time} to a very high value ($>= 5.0$) will result in long curves that are not very flexible. This problem is not that unavoidable,
because the planner actively replans after each time interval (controlled by \texttt{controller\_frequency}($Hz$)), which leaves room for small adjustments. A value of 4.0 seconds should be enough even for high performance computers.

\begin{figure}[!htb]
\minipage{0.49\textwidth}
  \includegraphics[width=\linewidth]{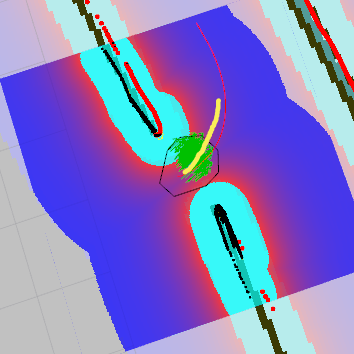}
  \caption{\texttt{sim\_time} = 1.5}
\endminipage\hfill
\minipage{0.49\textwidth}
  \includegraphics[width=\linewidth]{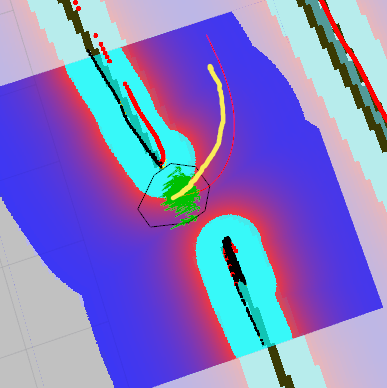}
  \caption{\texttt{sim\_time} = 4.0}
\endminipage\hfill
\end{figure}

Besides \texttt{sim\_time}, there are several other parameters that worth our attention.
\paragraph{Velocity samples}
Among other parameters, \texttt{vx\_sample}, \texttt{vy\_sample} determine how many translational velocity samples to take in x, y direction. \texttt{vth\_sample} controls the number of rotational velocities samples. The number of samples you would like to take depends on how much computation power you have. In most cases we prefer to set \texttt{vth\_samples} to
be higher than translational velocity samples, because turning is generally a more complicated condition than moving straight ahead. If you set \texttt{max\_vel\_y} to be zero,
there is no need to have velocity samples in y direction since there will be no usable samples. We picked \texttt{vx\_sample} = 20, and \texttt{vth\_samples} = 40.

\paragraph{Simulation granularity}
\texttt{sim\_granularity} is the step size to take between points on a trajectory. It basically means how frequent should the points on this trajectory be examined (test if they intersect with any obstacle or not). A lower value means higher frequency, which requires more computation power. The default value of 0.025 is generally enough for turtlebot-sized mobile base.

\subsubsection{DWA Local Planner : Trajactory Scoring}

As we mentioned above, DWA Local Planner maximizes an objective function to obtain optimal velocity pairs. In its paper, the value of this objective function relies on three
components: progress to goal, clearance from obstacles and forward velocity. In ROS's implementation, the cost of the objective function is calculated like this:

\footnotesize
\begin{eqnarray*}
    cost &=& \texttt{path\_distance\_bias} * (\text{distance($m$) to path from the endpoint of the trajectory})\\
    &+& \texttt{goal\_distance\_bias} * (\text{distance($m$) to local goal from the endpoint of the trajectory})\\
    &+& \texttt{occdist\_scale} * (\text{maximum obstacle cost along the trajectory in obstacle cost (0-254)})
\end{eqnarray*}

\normalsize
\noindent The objective is to get the lowest cost. \texttt{path\_distance\_bias} is the weight for how much the local planner should stay close to the global path \cite{furrer2016robot}. A high value will make the local planner prefer trajectories
on global path. 

\texttt{goal\_distance\_bias} is the weight for how much the robot should attempt to reach the local goal, with whatever path. Experiments show that increasing this
parameter enables the robot to be less attached to the global path.

\texttt{occdist\_scale} is the weight for how much the robot should attempt to avoid obstacles. A high value for this
parameter results in indecisive robot that stucks in place. Currently for SCITOS G5, we set \texttt{path\_distance\_bias} to 32.0, \texttt{goal\_distance\_bias} to 20.0, \texttt{occdist\_scale} to 0.02. They work well in simulation.

\newpage
\subsubsection{DWA Local Planner : Other Parameters}

\paragraph{Goal distance tolerance} These parameters are straightforward to understand. Here we will list their description shown on ROS Wiki:

\begin{itemize}
    \item \texttt{yaw\_goal\_tolerance} (double, default: 0.05)
    The tolerance in radians for the controller in yaw/rotation when achieving its goal.
    \item \texttt{xy\_goal\_tolerance} (double, default: 0.10)
    The tolerance in meters for the controller in the x \& y distance when achieving a goal.
    \item \texttt{latch\_xy\_goal\_tolerance} (bool, default: false) If goal tolerance is latched, if the robot ever reaches the goal xy location it will simply rotate in place, even if it ends up outside the goal tolerance while it is doing so.
\end{itemize}
    
\paragraph{Oscilation reset} In situations such as passing a doorway, the robot may oscilate back and forth because its local planner is producing paths leading to two opposite directions. If the robot keeps oscilating, the navigation stack will let the robot try its recovery behaviors.
\begin{itemize}
    \item \texttt{oscillation\_reset\_dist} (double, default: 0.05) How far the robot must travel in meters before oscillation flags are reset.
\end{itemize}

\section{Costmap Parameters}

As mentioned above, \textit{costmap} parameters tuning is essential for the success of local planners (not only for DWA). In ROS, costmap is composed of static map layer, obstacle map layer and inflation layer. Static map layer directly interprets the given static SLAM map provided to the navigation stack. Obstacle map layer includes 2D obstacles and 3D obstacles (voxel layer). Inflation layer is where obstacles are inflated to calculate cost for each 2D costmap cell.

Besides, there is a \textit{global costmap}, as well as a \textit{local costmap}. Global costmap is generated by inflating the obstacles on the map provided to the navigation stack. Local costmap is generated
by inflating obstacles detected by the robot's sensors in real time. \\

\begin{figure}[!h]
  \includegraphics[width=\linewidth]{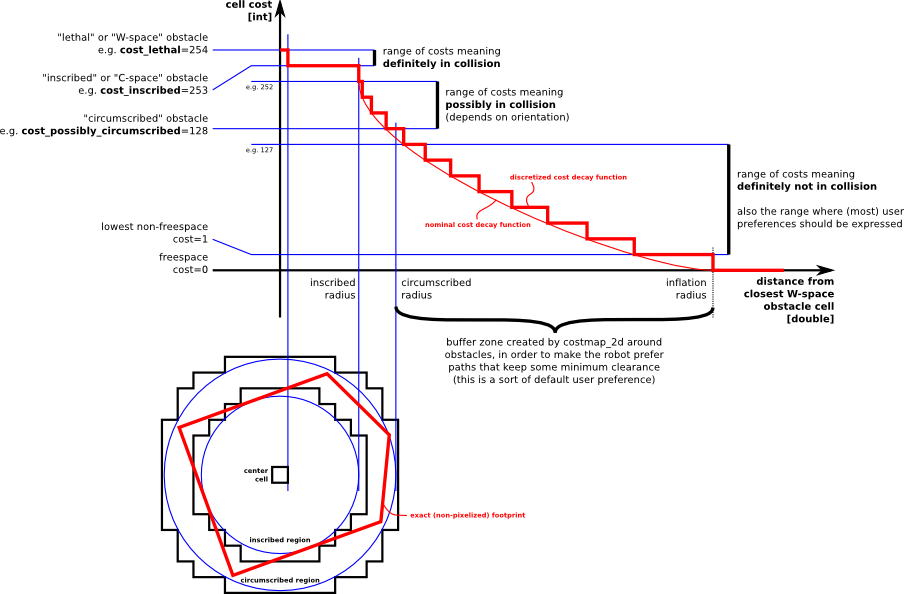}
  \caption{inflation decay}
\end{figure}

\noindent There are a number of important parameters that should be set as good as possible.

\subsection{footprint} 
Footprint is the contour of the mobile base. In ROS, it is represented by a two dimensional array of the form $[x_0, y_0],[x_1,y_1],[x_2,y_2],...]$, no need to repeat the first coordinate. This footprint will be used to compute the radius of inscribed circle and circumscribed circle, which are used to inflate obstacles in a way that fits this robot. Usually for safety, we want to have the footprint to be slightly larger
than the robot's real contour.

To determine the footprint of a robot, the most straightforward way is to refer to the drawings of your robot. Besides, you can manually take a picture of the top view of its base. Then use CAD software (such as Solidworks) to scale the image appropriately and move your mouse around the contour of the base and read its coordinate. The origin of the coordinates should be the center of the robot. Or, you can move your robot on a piece of large paper, then draw the contour of the base. Then pick some vertices and use rulers to figure out their coordinates.

\subsection{inflation} 
Inflation layer is consisted of cells with cost ranging from 0 to 255. Each cell is either occupied, free of obstacles, or unknown. Figure 13 shows a diagram \footnote{Diagram is from \url{http://wiki.ros.org/costmap_2d}} illustrating how inflation decay curve is computed.

\texttt{inflation\_radius} and \texttt{cost\_scaling\_factor} are the parameters that determine the inflation. \texttt{inflation\_radius} controls how far away the zero cost point is from the obstacle. \texttt{cost\_scaling\_factor} is inversely proportional to the cost of a cell. Setting it higher will make the decay curve more steep.

Dr. Pronobis sugggests the optimal costmap decay curve is one that has relatively low slope, so that the best path is as far as possible from the obstacles on each side. The advantage is that the robot would prefer to move in the middle of obstacles.  As shown in Figure 8 and 9, with the same starting point and goal, when costmap curve is steep, the robot tends to be close to obstacles. In Figure 14, \texttt{inflation\_radius} = 0.55, \texttt{cost\_scaling\_factor} = 5.0; In Figure 15, \texttt{inflation\_radius} = 1.75, \texttt{cost\_scaling\_factor} = 2.58

\begin{figure}[!htb]
\minipage{0.49\textwidth}
  \includegraphics[width=\linewidth]{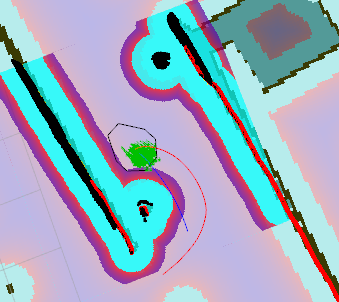}
  \caption{steep inflation curve}
\endminipage\hfill
\minipage{0.49\textwidth}
  \includegraphics[width=\linewidth]{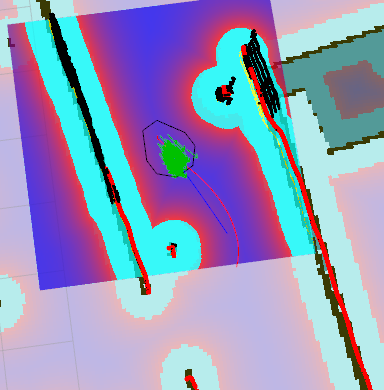}
  \caption{gentle inflation curve}
\endminipage\hfill
\end{figure}

Based on the decay curve diagram, we want to set these two parameters such that the inflation radius almost covers the corriders, and the decay of cost value is moderate, which means decrease the value of \texttt{cost\_scaling\_factor} .

\subsection{costmap resolution} 
This parameter can be set separately for local costmap and global costmap. They affect computation load and path planning. With low resolution ($>=0.05$), in narrow passways, the obstacle region may overlap and thus the local planner will not be able to find a path through.

For global costmap resolution, it is enough to keep it the same as the resolution of the map provided to navigation stack. If you have more than enough computation power,
you should take a look at the resolution of your laser scanner, because when creating the map using gmapping, if the laser scanner has lower resolution than your desired
map resolution, there will be a lot of small "unknown dots" because the laser scanner cannot cover that area, as in Figure 10.
\begin{figure}[!h]
    \begin{center}
        \includegraphics[width=22em]{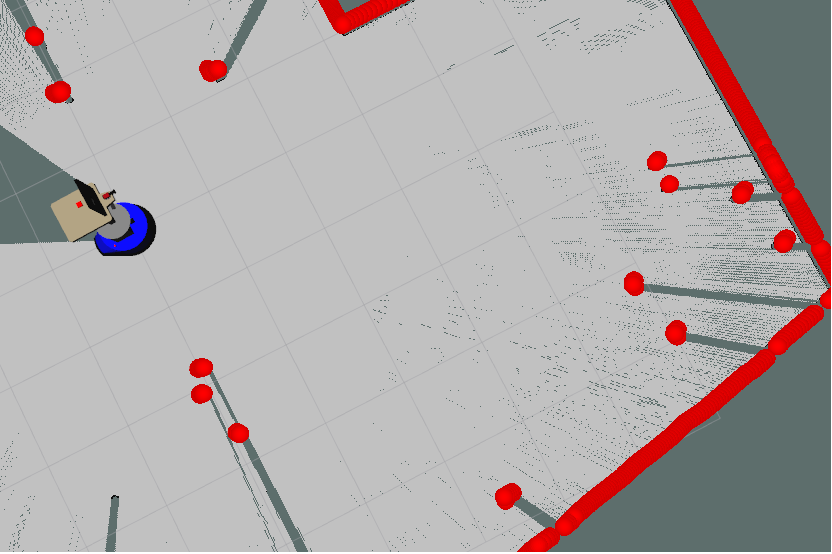}
      \caption{gmapping resolution = 0.01. Notice the unknown\\ dots on the right side of the image}
    \end{center}
\end{figure}

For example, Hokuyo URG-04LX-UG01 laser scanner has metric resolution of 0.01mm\footnote{data from \url{https://www.hokuyo-aut.jp/02sensor/07scanner/download/pdf/URG-04LX_UG01_spec_en.pdf}}. Therefore, scanning a map with resolution $<=0.01$ will require the robot to rotate several times in order to clear unknown dots. We found 0.02 to be a sufficient resolution
to use.

\subsection{obstacle layer and voxel layer} These two layers are responsible for marking obstacles on the costmap. They can be called altogether as \textit{obstacle layer}. According to ROS wiki, the obstacle layer tracks in two dimensions, whereas the voxel layer tracks in three. Obstacles are marked (detected) or cleared (removed) based on data from robot's sensors, which has topics for costmap to subscribe to.

In ROS implementation, the voxel layer inherits from obstacle layer, and they both obtain obstacles information by interpreting laser scans or data sent with \texttt{PointCloud} or \texttt{PointCloud2} type messages. Besides, the voxel layer requires depth sensors such as Microsoft Kinect or ASUS Xtion. 3D obstacles are eventually projected down to the 2D costmap for inflation.

\paragraph{How voxel layer works} Voxels are 3D volumetric cubes (think 3D pixel) which has certain relative position in space. It can be used to be associated with data or properties of the volume near it, e.g. whether its location is an obstacle. There has been quite a few research around online 3D reconstruction with the depth cameras via voxels. Here are some of them.

\begin{itemize}
    \item \href{http://delivery.acm.org/10.1145/2050000/2047270/p559-izadi.pdf?ip=128.208.7.188&id=2047270&acc=ACTIVE\%20SERVICE&key=B63ACEF81C6334F5\%2EF43F328D6C8418D0\%2E4D4702B0C3E38B35\%2E4D4702B0C3E38B35&CFID=830915711&CFTOKEN=23054788&__acm__=1472349664_9fd28ae246d72a507f6a93c5ac84a516}{KinectFusion: Real-time 3D Reconstruction and Interaction Using a Moving Depth Camera} 
    \item \href{https://people.mpi-inf.mpg.de/~mzollhoef/Papers/SGASIA2013_VH/paper.pdf}{Real-time 3D Reconstruction at Scale using Voxel Hashing}
\end{itemize}

\texttt{voxel\_grid} is a ROS package which provides an implementation of efficient 3D voxel grid data structure that stores voxels with three states: marked, free, unknown. The \textit{voxel grid} occupies the volume within the costmap region. During each update of the voxel layer's boundary, the voxel layer will mark or remove some of the voxels in the voxel grid based on observations from sensors. It also performs ray tracing, which is discussed next. Note that the voxel grid is not recreated when updating, but only updated unless the size of local costmap is changed.

\paragraph{Why ray tracing} in obstacle layer and voxel layer? Ray tracing is best known for rendering realistic 3D graphics, so it might be confusing why it is used in dealing with obstacles. One big reason is that obstacles of different type can be detected by robot's sensors. Take a look at figure 17. In theory, we are also able to know if an obstacle is rigid or soft (e.g. grass)\footnote{ mentioned in \textit{Using Robots in Hazardous Environments} by Boudoin, Habib, pp.370}.

\begin{figure}[!h]
    \begin{center}
        \includegraphics[width=23em]{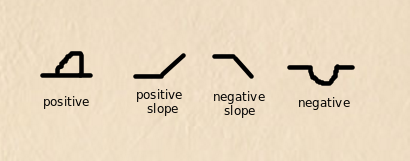}
      \caption{With ray tracing, laser scanners is able to recognize different types of obstacles.}
    \end{center}
\end{figure}

A good blog on voxel ray tracing versus polygong ray tracing: \url{http://raytracey.blogspot.com/2008/08/voxel-ray-tracing-vs-polygon-ray.html} \\

With the above understanding, let us look into the parameters for the obstacle layer\footnote{Some explanations are directly copied from costmap2d ROS Wiki}. These parameters are global filtering parameters that apply to all sensors.

\begin{itemize}
\item \texttt{max\_obstacle\_height}: The maximum height of any obstacle to be inserted into the costmap in meters. This parameter should be set to be slightly higher than the height of your robot. For voxel layer, this is basically the height of the voxel grid.

\item \texttt{obstacle\_range}: The default maximum distance from the robot at which an obstacle will be inserted into the cost map in meters. This can be over-ridden on a per-sensor basis.
\item \texttt{raytrace\_range}: The default range in meters at which to raytrace out obstacles from the map using sensor data. This can be over-ridden on a per-sensor basis.
\end{itemize}

These parameters are only used for the voxel layer (VoxelCostmapPlugin).

\begin{itemize}
\item \texttt{origin\_z}: The z origin of the map in meters.
\item \texttt{z\_resolution}: The z resolution of the map in meters/cell.
\item \texttt{z\_voxels}: The number of voxels to in each vertical column, the height of the grid is z\_resolution * z\_voxels.
\item \texttt{unknown\_threshold}: The number of unknown cells allowed in a column considered to be "known"
\item \texttt{mark\_threshold}: The maximum number of marked cells allowed in a column considered to be "free".
\end{itemize}

\paragraph{Experiment observations} Experiments further clarify the effects of the voxel layer's parameters.  We use ASUS Xtion Pro as our depth sensor. We found that position of Xtion matters in that it determines the range of "blind field", which is the region that the depth sensor cannot see anything. 

In addition, voxels representing obstacles only update (marked or cleared) when obstacles appear within Xtion range. Otherwise, some voxel information will remain, and their influence on costmap inflation remains. 

Besides, \texttt{z\_resolution} controls how dense the voxels is on the $z$-axis. If it is higher, the voxel layers are denser. If the value is too low (e.g. 0.01), all the voxels will be put together and thus you won't get useful costmap information. If you set z\_resolution to a higher value, your intention should be to obtain obstacles better, therefore you need to increase \texttt{z\_voxels} parameter which controls how many voxels in each vertical column. It is also useless if you have too many voxels in a column but not enough resolution, because each vertical column has a limit in height. Figure 18-20 shows comparison between different voxel layer parameters setting.

\begin{figure}[!h]
\end{figure}

\begin{figure}[!h]
\minipage{0.32\textwidth}
    \includegraphics[width=\linewidth]{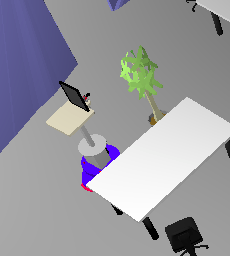}
  \caption{Scene: Plant in front of the robot}
\endminipage\hfill
\minipage{0.32\textwidth}
  \includegraphics[width=\linewidth]{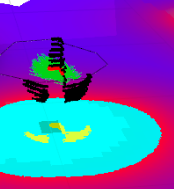}
  \caption{high \texttt{z\_resolution}}
\endminipage\hfill
\minipage{0.32\textwidth}
  \includegraphics[width=\linewidth]{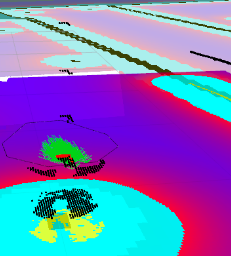}
  \caption{low \texttt{z\_resolution}}
\endminipage\hfill
\end{figure}

\section{AMCL}

\texttt{amcl} is a ROS package that deals with robot localization. It is the abbreviation of Adaptive Monte Carlo Localization (AMCL), also known as partical filter localization. This localization technique works like this: Each sample stores a position and orientation data representing the robot's pose. Particles are all sampled randomly initially. When the robot moves, particles are resampled based on their current state as well as robot's action using recursive Bayesian estimation.

More discussion on AMCL parameter tuning will be provided later. Please refer to \url{http://wiki.ros.org/amcl} for more information. For the details of the original algorithm Monte Carlo Localization (MCL), read Chapter 8 of \textit{Probabilistic Robotics} \cite{thrun2005probabilistic}.\\

\noindent We now summarize several issues that may affect the quality of AMCL localization\footnote{Added on April 8th, 2019. This investigation was done in May, 2017, yet not reported in this guide at the time.}.  We hope this information makes this guide more complete, and you find it useful.

\begin{figure}[!tb]
\minipage{0.49\textwidth}
    \includegraphics[width=\linewidth]{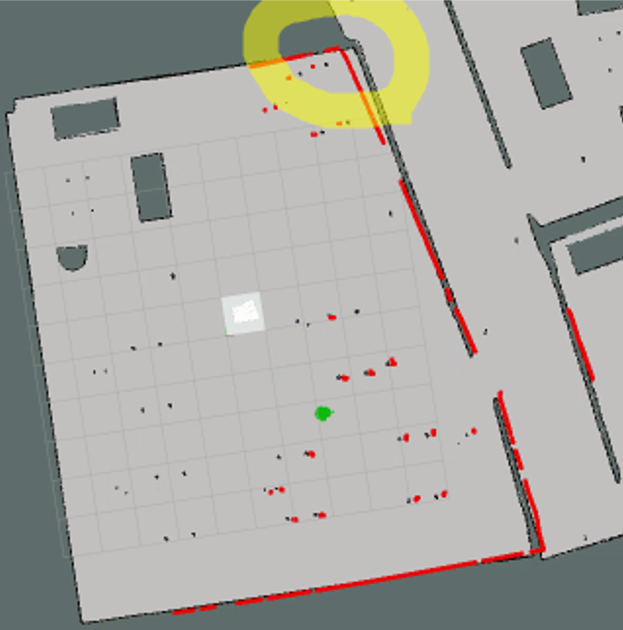}
    \caption{When \texttt{LaserScan} fields are not correct}
    \label{fig:scanner1}
\endminipage\hfill
\minipage{0.49\textwidth}
  \includegraphics[width=\linewidth]{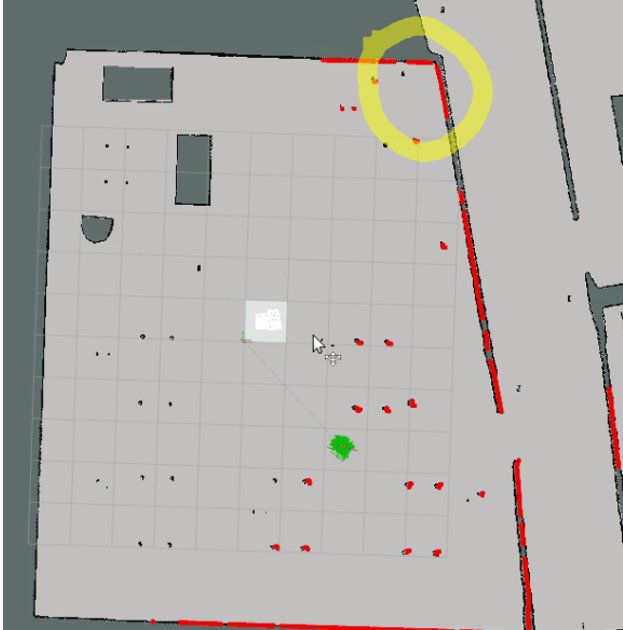}
  \caption{When \texttt{LaserScan} fields are correct}
  \label{fig:scanner2}
\endminipage\hfill
\end{figure}

Through experiments, we observed three issues that affect the localization with AMCL. As described in \cite{thrun2005probabilistic}, MCL maintains two probabilistic models, a \emph{motion model} and a \emph{measurement model}. In ROS \texttt{amcl}, the motion model corresponds to a model of the odometry, while the measurement model correspond to a model of laser scans. With this general understanding, we describe three issues separately as follows. 

\subsection{Header in \texttt{LaserScan} messages}

Messages that are published to \texttt{scan} topic are of type \texttt{sensor\_msgs/LaserScan}\footnote{See: \url{http://docs.ros.org/melodic/api/sensor_msgs/html/msg/LaserScan.html}}. This message contains a header with fields dependent on the specific laser scanner that you are using. These fields include (copied from the message documentation)

\begin{itemize}
\item \texttt{angle\_min} (float32) start angle of the scan [rad]
\item \texttt{angle\_max} (float32) end angle of the scan [rad]
\item \texttt{angle\_increment} (float32)  start angle of the scan [rad]
\item \texttt{time\_increment} (float32) time between measurements [seconds] - if your scanner is moving, this will be used in interpolating position of 3d points
\item \texttt{scan\_time} (float32) time between scans [seconds]
\item \texttt{range\_min} (float32) minimum range value [m]
\item \texttt{range\_max} (float32) maximum range value [m]
\end{itemize}

We observed in our experiments that if these values are not set correctly for the laser scanner product on board, the quality of localization will be affected (see Figure~\ref{fig:scanner1} and \ref{fig:scanner2}. We have used two laser scanners products, the SICK LMS 200 and the SICK LMS 291. We provide their parameters below\footnote{For LMS 200, thanks to {this Github issue} (\url{https://github.com/smichaud/lidar-snowfall/issues/1})}.

\noindent SICK LMS 200:
\lstset{language=json}
\begin{lstlisting}
{   
    "range_min": 0.0,
    "range_max": 81.0,
    "angle_min": -1.57079637051,
    "angle_max": 1.57079637051,
    "angle_increment": 0.0174532923847,
    "time_increment": 3.70370362361e-05,
    "scan_time": 0.0133333336562
}
\end{lstlisting}

\noindent SICK LMS 291:
\lstset{language=json}
\begin{lstlisting}
{   
    "range_min": 0.0,
    "range_max": 81.0,
    "angle_min": -1.57079637051,
    "angle_max": 1.57079637051,
    "angle_increment": 0.00872664619235,
    "time_increment": 7.40740724722e-05,
    "scan_time": 0.0133333336562
}
\end{lstlisting}

\subsection{Parameters for measurement and motion models}
There are parameters listed in the \texttt{amcl} package about tuning the laser scanner model (measurement)  and odometry model (motion). Refer to the package page for the complete list and their definitions. A detailed discussion requires great understanding of the MCL algorithm in \cite{thrun2005probabilistic}, which we omit here. We provide an \emph{example} of fine tuning these parameters and describe their results qualitatively. The actual parameters you use should depend on your laser scanner and robot.

For \textbf{laser scanner model}, the default parameters are:

\lstset{language=json}
\begin{lstlisting}
{
    "laser_z_hit": 0.5,
    "laser_sigma_hit": 0.2,
    "laser_z_rand" :0.5,
    "laser_likelihood_max_dist": 2.0
}
\end{lstlisting}

\begin{figure}[!tb]
\minipage{0.49\textwidth}
    \includegraphics[width=\linewidth]{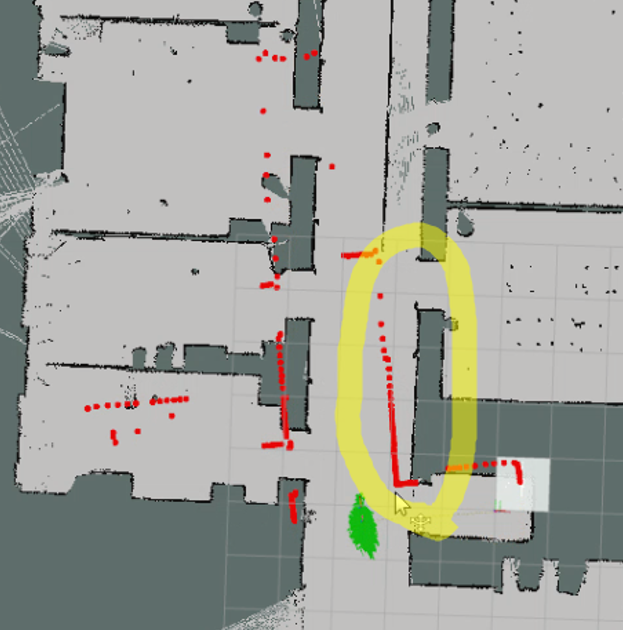}
    \caption{Default measurement model parameters}
    \label{fig:param1}
\endminipage\hfill
\minipage{0.49\textwidth}
  \includegraphics[width=\linewidth]{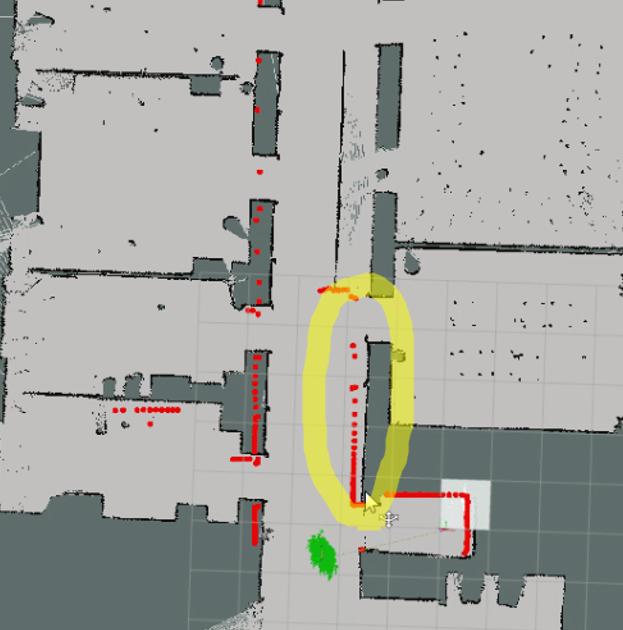}
  \caption{After tuning measurement model parameters (increase noise)}
  \label{fig:param2}
\endminipage\hfill
\end{figure}

To improve the localization of our robot, we increased \texttt{laser\_z\_hit} and \texttt{laser\_sigma\_hit} to incorporate higher measurement noise. The resulting parameters are:

\lstset{language=json}
\begin{lstlisting}
{
    "laser_z_hit": 0.9,
    "laser_sigma_hit": 0.1,
    "laser_z_rand" :0.5,
    "laser_likelihood_max_dist": 4.0
}
\end{lstlisting}
The behavior is illustrated in Figure~\ref{fig:param1} and \ref{fig:param2}. It is clear that in our case, adding noise into the measurement model helped with localization.

For the \textbf{odometry model}, we found that our odometry was quite reliable in terms of stability. Therefore, we tuned the parameters so that the algorithm assumes there is low noise in odometry:
\lstset{language=json}
\begin{lstlisting}
{   
    "kld_err": 0.01,
    "kld_z": 0.99,
    "odom_alpha1": 0.005,
    "odom_alpha2": 0.005,
    "odom_alpha3": 0.005,
    "odom_alpha4": 0.005
}
\end{lstlisting}
To verify that the above paremeters for motion model work, we also tried a set of parameters that suggest a noisy odometry model:
\lstset{language=json}
\begin{lstlisting}
{
    "kld_err": 0.10,
    "kld_z": 0.5',                                                                            
    "odom_alpha1": 0.008,                                                                     
    "odom_alpha2": 0.040,                                                                     
    "odom_alpha3": 0.004,                                                                     
    "odom_alpha4": 0.025                                                                      
} 
\end{lstlisting}
We observed that when the odometry model is less noisy, the particles are more condensed. Otherwise, the particles are more spread-out.

\subsection{Translation of the laser scanner}

There is a \texttt{tf} transform from \texttt{laser\_link} to \texttt{base\_footprint} or \texttt{base\_link} that indicates the pose of the laser scanner with respect to the robot base. If this transform is not correct, it is very likely that the localization behaves strangely. In this situation, we have observed constant shifting of laser readings from the walls of the environment, and sudden drastic change in the localization. It is straightforward enough to make sure the transform is correct; This is usually handled in \texttt{urdf} and \texttt{srdf} specification of your robot. However, if you are using a \texttt{rosbag} file, you may have to publish the transform youself.

\section{Recovery Behaviors}
An annoying thing about robot navigation is that the robot may get stuck. Fortunately, the navigation stack has recovery behaviors built-in. Even so, sometimes the robot will exhaust all available recovery behaviors and stay still. Therefore, we may need to figure out a more robust solution.

\paragraph{Types of recovery behaviors} ROS navigation has two recovery behaviors. They are \texttt{clear\_costmap\_recovery} and \texttt{rotate\_recovery}. Clear costmap recovery is basically reverting the local costmap to have the same state as the global costmap. Rotate recovery is to recover by
rotating 360 degrees in place. 

\paragraph{Unstuck the robot} Sometimes rotate recovery will fail to execute due to rotation failure. At this point, the robot may just give up because it has tried
all of its recovery behaviors - clear costmap and rotation. In most experiments we observed that when the robot gives up, there are actually many ways to unstuck the robot.
To avoid giving up, we used SMACH to continuously try different recovery behaviors, with additional ones such as setting a temporary goal that is very close to the robot, and returning to some previously visited pose (i.e. backing off). These methods turn out to improve the robot's durability substantially, and unstuck it from previously hopeless tight spaces from our experiment observations\footnote{Here is a video demo of my work on mobile robot navigation: \url{https://youtu.be/1-7GNtR6gVk}}.

\begin{figure}[!h]
    \begin{center}
        \includegraphics[width=26em]{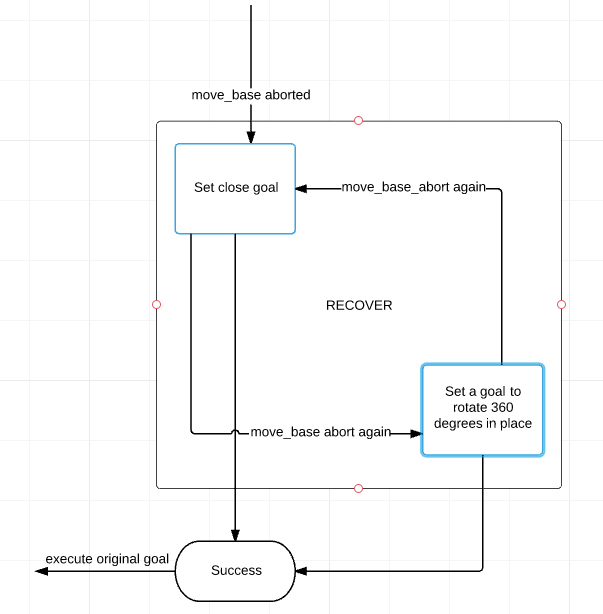} \caption{Simple recovery state in SMACH}
    \end{center}
\end{figure}

\paragraph{Parameters} The parameters for ROS's recovery behavior can be left as default in general. For clear costmap recovery, if you have a relatively
high \texttt{sim\_time}, which means the trajectory is long, you may want to consider increasing \texttt{reset\_distance} parameter, so that bigger area on local costmap is removed, and there is a
better chance for the local planner to find a path.

\section{Dynamic Reconfigure}

One of the most flexible aspect about ROS navigation is dynamic reconfiguration, since different parameter setup might be more helpful for certain situations, e.g. when robot is close to the goal. Yet, it is not necessary to do a lot of dynamic reconfiguration. 

\paragraph{Example} One situation that we observed in our experiments is that the robot tends to fall off the global path,
even when it is not necessary or bad to do so. Therefore we increased \texttt{path\_distance\_bias}. Since a high \texttt{path\_distance\_bias} will make the robot stick to the global path, which does not actually lead to the final goal due to tolerance, we need a way to let the robot reach the goal with no hesitation. We chose to dynamically
decrease the \texttt{path\_distance\_bias} so that \texttt{goal\_distance\_bias} is emphasized when the robot is close to the goal. After all, doing more experiments is the ultimate way to find problems and figure out solutions.

\section{Problems}

\begin{enumerate}
\item Getting stuck

This is a problem that we face a lot when using ROS navigation. In both simulation and reality, the robot gets stuck and gives up the goal. 

\item Different speed in different directions

We observed some weird behavior of the navigation stack. When the goal is set in the -x direction with respect to TF origin, dwa local planner plans less stably (the local planned trajectory jumps around) and the robot moves really slowly. But when the goal is set in the +x direction, dwa local planner is much more stable, and the robot can move faster.

I reported this issue on Github here: \url{https://github.com/ros-planning/navigation/issues/503}. Nobody attempted to resolve it yet.

\item Reality VS. simulation

There is a difference between reality and simulation. In reality, there are more obstacles with various shapes. For exmaple, in the lab there is a vertical stick that is used to hold to door
open. Because it is too thin, the robot sometimes fails to detect it and hit on it. There are also more complicated human activity in reality.

\item Inconsistency

Robots using ROS navigation stack can exhibit inconsistent behaviors, for example when entering a door, the local costmap is generated again and again with slight difference each time, and this may affect path planning, especially when resolution is low. Also, there is no memory for the robot. It does not remember how it entered the room from the door the last time. So it needs to start out fresh again every time it tries to enter a door. Thus, if it enters the door in a different angle than before, it may just get stuck and give up.

\end{enumerate}

\section*{Thanks}

Hope this guide is helpful. Please feel free to add more information from your own experimental observations.

\nocite{*}
\bibliographystyle{apalike}
\bibliography{main}

\end{document}